\documentclass[journal]{IEEEtran}
\usepackage{graphicx}
\usepackage{amsmath}
\usepackage{amssymb}
\usepackage{array}
\usepackage{xcolor}
\usepackage{cite}

\usepackage{booktabs}
\usepackage{multirow}
\usepackage{caption}
\usepackage{graphicx}
\usepackage{bm}
\usepackage{comment}
\usepackage{longtable}
\usepackage{tabularx}
\renewcommand{\arraystretch}{1.2}
\usepackage{algorithm}
\usepackage{algpseudocode} 
\usepackage{amsmath}
\usepackage{amssymb}
\title{algorithms}

\usepackage{subfigure} 
\usepackage{subcaption}

\usepackage{ifthen}
\usepackage{hyperref}

\usepackage[table]{xcolor}

\ifCLASSINFOpdf
\else
\fi

\begin{document}
%
\title{Joint Sensor Deployment and Physics-Informed Graph Transformer for Smart Grid Attack Detection}
%
%
%

\author{Mariam Elnour \textsuperscript{1}, 
 Mohammad AlShaikh Saleh\textsuperscript{2}, 
 Rachad Atat\textsuperscript{3}, 
 Xiang Huo\textsuperscript{4}, 
 Abdulrahman Takiddin\textsuperscript{5}, \\
 Muhammad Ismail\textsuperscript{6}, 
 Hasan Kurban\textsuperscript{7}, 
 Katherine R. Davis\textsuperscript{1}, 
 Erchin Serpedin\textsuperscript{1}
\thanks{\textsuperscript{1} M. Elnour, K. R. Davis, and E. Serpedin are with the department of Electrical and Computer Engineering, Texas A\&M University, College Station, TX 77840, USA. (email: melnour@tamu.edu, katedavis@tamu.edu, eserpedin@tamu.edu)}
\thanks{\textsuperscript{2} M. A. Saleh is with the department of Electrical and Computer Engineering, Texas A\&M University at Qatar, Doha, Qatar. (email: mohammad.saleh@qatar.tamu.edu)}
\thanks{\textsuperscript{3} R. Atat is with the department of Computer Science and Mathematics, Lebanese American University, Beirut 1102-2801, Lebanon. (email: rachad.atat@lau.edu.lb)}
\thanks{\textsuperscript{4} X. Huo is with the department of Electrical and Computer Engineering, Hampton University, 100 E Queen St, Hampton, VA 23669, USA (email: xiang.huo@hamptonu.edu) }
\thanks{\textsuperscript{5} A. Takiddin is with the department of Electrical and Computer Engineering, FAMU-FSU College of Engineering, Florida State University, 2525 Pottsdamer Street. Tallahassee, FL 32310-6046, USA. (email: a.takiddin@fsu.edu) }
\thanks{\textsuperscript{6} M. Ismail is with the Cybersecurity Education, Research, and Outreach Center (CEROC) and the department of Computer Science, Tennessee Tech University, Cookeville, TN, USA. (email: mismail@tntech.edu)}
\thanks{\textsuperscript{7} H. Kurban is with the College of Science and Engineering, Hamad Bin Khalifa University, Doha 34110, Qatar. (email: hkurban@hbku.edu.qa)}
}

\maketitle

\begin{abstract}
This paper proposes a joint multi-objective optimization framework for strategic sensor placement in power systems to enhance attack detection. A novel physics-informed graph transformer network (PIGTN)-based detection model is proposed. Non-dominated sorting genetic algorithm-II (NSGA-II) jointly optimizes sensor locations and the PIGTN’s detection performance, while considering practical constraints. The combinatorial space of feasible sensor placements is explored using NSGA-II, while concurrently training the proposed detector in a closed-loop setting. Compared to baseline sensor placement methods, the proposed framework consistently demonstrates robustness under sensor failures and improvements in detection performance in seven benchmark cases, including the 14, 30, IEEE-30, 39, 57, 118 and the 200 bus systems. By incorporating AC power flow constraints, the proposed PIGTN-based detection model generalizes well to unseen attacks and outperforms other graph network-based variants (topology-aware models), achieving improvements up to 37\% in accuracy and 73\% in detection rate, with a mean false alarms rate of 0.3\%. In addition, optimized sensor layouts significantly improve the performance of power system state estimation, achieving a 61\%--98\% reduction in the average state error.
\end{abstract}


\IEEEpeerreviewmaketitle

\section{Introduction}

{Power systems are becoming increasingly vulnerable to cyber-physical threats due to the rapid integration of new emerging smart grid technologies, such as advanced metering infrastructure, monitoring devices, microgrids, electric vehicles and charging units, cyber-physical energy systems and decentralization of energy trading and market. Undiscovered system vulnerabilities, if exploited by adversaries, can have a significant impact, such as the famous Ukrainian power grid blackout that took place in 2015 \cite{DoE}.} However, when used effectively and combined with artificial intelligence (AI), the sensor infrastructure can be used to enhance monitoring and defense strategies \cite{9453402}. 
Using historical sensor data, numerous data-driven attack detection methods were studied \cite{9097420}.

Specifically, most recent works used graph  neural networks (GNNs) extensively for attack detection in power systems and demonstrated notable performance. For instance, a spatio-temporal wavelet graph convolutional neural network (GCN) was presented in \cite{QU2024122736} to localize dummy false data injection attacks (FDIAs). In \cite{9582826}, spectral GCNs were proposed to support bad data detectors (BDDs) in stealthy FDIA detection.  
For greater flexibility and representational power, graph transformer-based models represent promising solutions, motivated by the outstanding performance observed of transformers in natural language processing \cite{vaswani2017attention} and computer vision \cite{shehzad2024graph}, along with the demonstrated value of GNNs as topology-aware models \cite{yun2019graph}. Graph transformers incorporate graph inductive bias (prior knowledge or assumptions about graph properties) to effectively process graph data \cite{shehzad2024graph}. They possess the capability to handle dynamic and heterogeneous graphs, and to adapt to changing network structures. 

Nevertheless, despite the remarkable results acquired by graph-based models, they are susceptible to poor extrapolation capabilities, scalability constraints, and the explainability of the system’s physical laws. Without embedding these laws, model predictions are mostly black-box outputs. Such limitations reduce their reliability as power system operators often face problems in interpreting their implications. To avoid such problems, physics-informed neural networks (PINNs) were introduced, integrating the fundamental physical principles of systems into neural network (NN) models \cite{raissi2019physics}. Extending this idea to graph transformer networks (GTNs), they can be enhanced to capture long-range dependencies with improved interpretability by  integrating physics-informed design. However, the reliability of such  models is still contingent on the quality and scope of their input data. 

This highlights a critical dependency between modeling and sensing. Even for advanced models such as physics-informed graph transformer networks (PIGTNs), their performance is fundamentally bounded by the representativeness of the data used in model training and operation. Existing work using data-driven detection methods in power systems often assumes that sensors are available at all nodes, overlooking the significant impact of sensor placement on detection efficacy. This assumption simplifies modeling and does not reflect real-world scenarios where sensor deployment may be less extensive due to budgetary, logistic, and other practical constraints.  
This dual perspective motivates this work where  PIGTNs provide a principled means to model and detect complex attacks with improved reliability, while joint sensor deployment ensures that the model receives representative and physically meaningful data to operate effectively. 

\subsection{PINNs in Power Systems}

PINNs, a branch of scientific machine learning, combined data-driven and physics-driven approaches to solve science and engineering problems \cite{raissi2019physics,MCCLENNY2023111722}. These PINN models utilized prior knowledge of physics through algebraic (less common) and differential equations that explain the relationship among the variables in the given data. The role of PINNs in cyber attack detection in  power systems is still emerging. According to \cite{ReviewPINN}, PINNs were used for state/parameter estimation, dynamic analysis, power flow calculation/voltage prediction, optimal power flow applications, and fault diagnosis. However, no studies have yet comprehensively researched PINN-based cyber-attack detection, especially in the case of different types of attacks imposed on various network topologies. In \cite{Zideh2024}, a physics-informed convolutional autoencoder (PIConvAE) was applied to detect FDIAs in power distribution grids. Although promising performance was achieved, the study lacked comprehensive benchmark evaluations as it was evaluated on two test cases, the IEEE 13-bus and 123-bus systems, and against a single benchmark detector.  

To address this gap, this paper presents a comprehensive evaluation of the PIGTN--a physics-informed graph-aware data-driven model, that integrates the physics-based relationships of phasor measurements into the graph-based detection model. To the best of the authors' knowledge, we propose the first PIGTN-based cyber-attack detector for power systems. The proposed PIGTN leverages the graph transformer network (GTN) to process graph-structured data of the electrical topologies while enhancing explainability due to encoding the physical laws, i.e., the AC power flow equations, as a regularization term in the loss function, resulting in physically consistent predictions.

\begin{table*}[htp]
\centering
\scriptsize
\renewcommand{\arraystretch}{1.3}
\caption{Summary of related works in sensor placement for power systems.}
\label{tab:optimization-summary}
\resizebox{\linewidth}{!}{
\begin{tabular}{
    >{\raggedright\arraybackslash}p{0.5cm}
    >{\raggedright\arraybackslash}p{2.3cm}
    >{\raggedright\arraybackslash}p{3cm}
    >{\raggedright\arraybackslash}p{1.8cm}
    >{\raggedright\arraybackslash}p{5.2cm}
    >{\raggedright\arraybackslash}p{3.5cm}
    >{\raggedright\arraybackslash}p{5cm}
}
\toprule
\textbf{Ref} & \textbf{Task} & \textbf{Diagnosis Model} & \textbf{Optimization} & \textbf{Objective Function} & \textbf{Constraints} & \textbf{Joint or Standalone?} \\
\midrule
\cite{10640266} & Line fault detection  & Model- and Data-based & ILP & Min. number of PMUs & Coverage & Joint (model-based), Standalone (data-driven) \\
\cite{8765426} & Line fault detection & Model-based & ILP & Min. number of PMUs & Coverage, Redundancy & Standalone \\
\cite{8892526} & Line fault localization & Model-based & Greedy & Min. influence of unmeasured on measured buses & N/A & Joint to increase fault localization resolution \\
\cite{10703077} & Fault diagnosis & Data-driven & ILP & Min. number of PMUs & Topological observability & Standalone \\
\cite{8082533} & Anomaly detection & Rule-based & Min–Max & Min. max eigenvalue of sensitivity matrix & Placement feasibility & Standalone \\
\cite{9031416} & FDIA detection & Model-based & Greedy & Min. number of PMUs & Numerical observability & Joint with state estimation \\
\cite{10038443} & FDIA detection & Model-based & Greedy & Min. projection energy in normal subspace & N/A & Joint to improve SPD-based detection \\
\cite{7885045} & FDIA defense & Model-based & Greedy & Max. resilience of state estimation & N/A & Joint for stealthy FDIA defense \\
\bottomrule
\end{tabular}}
\end{table*}

\subsection{Sensor Placement in Power Systems}

Traditionally, the sensor placement problem was studied in the context of system observability for power system state estimation (PSSE) and energy management system (EMS) functions \cite{6135526, 6777591, 9774839}. Early works employed optimization techniques including integer linear programming  (ILP) \cite{4839066,4555699, 8827580}, 
greedy algorithms \cite{4519389}, and evolutionary algorithms \cite{10938564, 10018497} to ensure minimal-cost deployments.  
They focused almost exclusively on achieving topological or numerical observability. However, adversaries could craft attacks that precisely exploit this fully observable structure to remain hidden from conventional detection schemes. Additionally, when measurements are concentrated on  specific buses or are unevenly spread across the network, attackers may leverage this setting to manipulate poorly monitored subsets of the available sensor network and achieve their goal. Consequently, a placement strategy that satisfies observability alone may not guarantee attack--resilience, a gap that is not addressed enough in the literature.

Recent works explored optimal sensor placement for fault or anomaly detection (Table \ref{tab:optimization-summary}).  For instance, a strategic phasor measurement unit (PMU) placement strategy using ILP was proposed in \cite{10640266} to maximize the performance of model-based sparse estimation-based line outage and fault diagnosis. The study in \cite{8765426} focused on PMUs placement for wide-area backup protection such that every line fault was detectable based on voltage path differences. In \cite{8892526},  a PMU placement strategy was presented for a near-optimal resolution to localize faults given the limitations of available PMUs using graph signal sampling.  
An ILP sensor placement problem was reported in \cite{10703077} that aimed to minimize the number of PMUs,  incorporating  phase observability constraints  
to ensure components needed for diagnosis were measured. Then a neural network was developed for the diagnosis of the fault followed by an ensemble algorithm for the precise localization of the line level.  In \cite{8082533},  a hierarchical anomaly detection framework for distribution grids was proposed, combining optimally placed PMUs with physics-informed local and central detection rules to identify deviations from quasi-steady-state operation.  These strategies improved fault and anomaly detection; however, such anomalies are not necessarily cyber-attacks, and attacks can remain undetected by methods designed only for faults or random anomalies \cite{6032057}. Additionally, they were mainly model-based, and for the data-driven methods, the sensor placement was optimized separately  from the detection model. 

A second category of works optimized sensor placement specifically to improve attack detection. The authors in \cite{10038443} studied the performance of FDIA subspace projection detection (SPD) given the problem of PMU placement. The optimization aimed to place PMUs in order to minimize the energy retained in the normal (healthy) subspace. 
In \cite{7885045}, a greedy PMU placement strategy was proposed to improve the performance of PSSE, by deploying PMUs that maximally protect the most vulnerable measurements.   A pre‑deployment PMU-based greedy algorithm was proposed in \cite{9031416} that aimed first to protect attack-vulnerable buses, then iteratively added further PMUs until numerical observability was achieved. These methods present a critical limitation: they are model-based, which may be ineffective against sophisticated attacks that align with the models used.   

Notably, none integrated the placement optimization directly with data-driven detection models.  Our proposed framework bridges this gap by jointly optimizing sensor placement and detection performance. It aims to explicitly address the interplay between data-driven detection models, specifically graph-based, and sensor network in power systems. It addresses the fundamental problem of detection capability of graph learning models in the presence of practical constraints.

\subsection{Contributions}
We propose a joint optimization framework that simultaneously considers sensor placement and attack detection. It integrates both in a single, multi-objective optimization process instead of treating these tasks in isolation. We employ Genetic Algorithm (GA) to explore the combinatorial space of feasible sensor placements under relevant constraints \cite{10664002}, and concurrently train the attack detection model in a closed-loop setting. The framework solves for an optimized sensor layout that ensures an effective trade-off between detection performance and sensor placement objectives. 

{In summary, our contributions are as follows:}
\begin{itemize}
    \item We address the connection between sensor placement and data-driven attack detection in power systems by jointly optimizing these two problems. The proposed framework combines constrained GA-based sensor placement with supervised attack PIGTN-based detection. We propose a hybrid node importance score that integrates both topological and electrical characteristics of the power network. This score guides the GA initialization and mutation processes to promote a more effective exploration–exploitation balance. It results in optimized sensor layouts for improved overall detection performance.    
    \item We incorporate governing physical laws (AC power flow equations) into the proposed model's loss function, guiding the predictions to remain consistent with the real-world constraints of the system rather than relying merely on statistical correlations in the data. {Our benchmarking evaluation considers seven other robust topology-aware detectors using  14, 30, IEEE-30, 39, 57, 118, and 200 bus systems.} The proposed framework shows strong generalization to unseen attacks during testing. Compared to benchmark detectors, our PIGTN consistently achieves superior performance. On average, the proposed PIGTN outperforms the best-performing baselines with an average F1-score of 93\% across all cases, and a mean FPR of 2\%, providing performance improvements ranging from 2.23\%--27.5\% (averaged across all seven bus system cases) relative to the state-of-the-art topology-aware graph models.
    \item Our proposed framework consistently outperforms benchmark placement methods across the seven test cases. Detection rate improvements are case-dependent, ranging from mixed to around 7\%. The varied improvements reflect the influence of network functional and structural characteristics, while false alarm rates are maintained at comparable or lower levels. More importantly, under sensor failures, the framework yields the most robust sensor layouts. Specifically, GA-PIGTN achieves higher robustness scores (up to 0.907),  and superior average accuracy (4.8\% improvement vs. 3\% for benchmarks), F1-score (2.6\% vs. 1.3--1.6\%), and precision (6.5\% vs. 3.2--3.7\%) across varied sensor failure levels. 
    \item In addition to enhancing attack detection, the optimized sensor layout leads to improved PSSE performance in all test cases, with  61\% to 94\% reduction in voltage magnitude error and 77\% to 98\% reduction in voltage angle error, evaluated on the weighted least squares (WLS) state estimator.
\end{itemize}

The rest of the paper is organized as follows. Section \ref{attack_model}  outlines attack models considered in this work. The problem formulation is presented in Section \ref{problem_form}, and the details of the optimization methodology and proposed PIGTN detector are presented in Section \ref{proposed_work}. Section \ref{res-discuss} presents the experimental results,  performance comparisons, and discussion. Finally, Section \ref{conc} concludes the paper and suggests directions for future research.

\section{Attack Modeling}
\label{attack_model}
The attack modeling focuses on FDIAs, where an adversary strategically manipulates or injects false data into the system to compromise its integrity, accuracy, or functionality. It is a critical attack as it can result in mislead monitoring, decision-making, or control processes \cite{deng2016false, Farag2020, HUSNOO2023344}.  
This class of FDIAs are crafted from normal measurement data \cite{10016905}.
\subsubsection{Random Attack}
The  measurement data is perturbed slightly. The alteration is modeled as a small perturbation factor \(\alpha\), which is applied uniformly across the benign samples. The adversarial measurement \(Z_s(t,i)\) at timestamp \(t\) and bus \(i\) is generated by \cite{10016905}: 
 \(Z_s(t,i) = (1 + \alpha)Z_b(t,i),\)
    where \(Z_b(t,i)\) is the original benign measurement. 
\subsubsection{General Attack}
    It introduces larger, more systematic changes to the  measurement data. It manipulates the data according to \cite{10016905}:  
 \( Z_s(t,i) = Z_b(t,i) + (-1)^\beta \alpha \: \gamma \: \text{Range}(Z_b(i)),\)
    where \(\beta\) is a binary random variable, \(\alpha\) is the attack magnitude,  \(\gamma\) is a uniform random variable between \(0\) and \(1\), and \(\text{Range}(.)\) denotes the range of the measurement.
    
    \subsubsection{Load Redistribution (LR) Attacks} 
    They are a class of cyber-attacks that redistribute loads among the buses while keeping the net load unchanged. The false loads in an LR attack satisfy \cite{chu2020detecting, 10190177}:
\(Z_s(t,i) = Z_b(t,i) + \Delta Z(t,i), \quad \sum_i \Delta Z(t,i) = 0,\)
where \(\Delta {Z}\) is the load change caused by the attack. The \textit{load shift} \(\tau\) quantifies the attack's detectability and is defined as the largest load change in percentage of the true loads:
\(\tau = \max_i \left| \frac{\Delta Z(t,i)}{Z_b(t,i)} \right| \times 100\%.\)
Attacks with large \(\tau\) are easily detectable, so adversaries' typical limit is \(\tau \leq 20\%\).

\section{Problem Formulation}
\label{problem_form}
The sensor placement problem is formulated as a constrained multi-objective optimization problem. The goal is to minimize the number of installed sensors and to simultaneously enhance graph-based attack detection performance. Non-dominated sorting genetic algorithm II (NSGA-II) is used because it maintains solution diversity and effectively balances competing objectives \cite{996017}.  The formal optimization formulation is described in the following subsections.

\subsubsection{Decision Variables}
The decision variable is the binary vector, \(
X = [x_1, x_2, \dots, x_{N_c}],\quad x_i \in \{0,1\}\), where $N_c$ is the number of candidate nodes for sensor placement, and $x_i = 1$ indicates a sensor is placed  at node $i$, and $x_i = 0$ otherwise. 

\subsubsection{Objectives}
The multi-objective optimization problem is formulated as:
\begin{equation}
\begin{split} 
&\text{min}_{X}  \left (f_1(X) + f_2(X) \right) \\
&\text{\textit{s.t.}} \quad V \leq \varepsilon,
\end{split}
\end{equation}
where \(f_1(X) = \sum_{i=1}^{N_c} x_i\) is the number of sensors \cite{10664002} and  
    \(f_2(X) = \mathcal{L}\left(y_k, \hat{y}_k\right)\) is the detection performance cost, with      $\mathcal{L}(.)$ as the model loss function, $y_k$ as the true labels, and $\hat{y}_k$ as the predicted labels. The optimization incorporates requirements related to sensor layout's connectivity, coverage, and redundancy. A total constraints violation ($V$) is defined as:  \(V
 \;=\; w_{\text{cn}}  P_{\text{connectivity}} + w_{\text{cr}} P_{\text{coverage}} + w_{\text{rd}} P_{\text{redundancy}},\)  representing the weighted aggregated violations of individual constraints, where \(w_{\text{cr}}, \, w_{\text{rd}}, \,w_{\text{rd}}\) denote connectivity, coverage, and redundancy violation weights, respectively, and $\varepsilon$ stands for the constraints violation tolerance. For a power network modeled as a graph \( \mathcal{G} = (\mathcal{V}, \mathcal{E}) \), where \( \mathcal{V} \) indicates the buses and \( \mathcal{E} \) represents the transmission lines or transformers, the requirements are the following: 
\paragraph{Connectivity}
Let $\mathcal{S} = \{i \in \mathcal{V} \, | \, x_i = 1\}$ be the set of sensor-equipped nodes and $\mathcal{E}_{\mathcal{S}}$ be the active edges. It is desirable for the sensor network to form a connected graph, $\mathcal{G}_{\mathcal{S}}$, to provide the graph-based detection model with a well-connected receptive field message passing \cite{khemani2024review}. A virtual flow-based formulation is used (Eq. \eqref{conn_const}), where a single flow is initiated at a selected root node $k \in \mathcal{S}$ and must reach all other selected nodes through activated edges. The connectivity constraints are defined as:
\begin{equation}
\label{conn_const}
\begin{split}
\sum_{j \in \mathcal{N}(k)} f_{kj} &= \sum_{\substack{i \in \mathcal{V} \\ i \neq k}} x_i \\
\sum_{j \in \mathcal{N}(i)} f_{ji} - \sum_{j \in \mathcal{N}(i)} f_{ij} &= -x_i, \quad \forall i \in \mathcal{V} \setminus \{k\}  \\
0 \leq f_{ij} &\leq (|\mathcal{V}| - 1) \cdot y_{ij}, \quad \forall (i,j) \in \mathcal{E} \\
y_{ij} &\leq x_i, \quad y_{ij} \leq x_j, \quad \forall (i,j) \in \mathcal{E}
\end{split}
\end{equation}
where $f_{ij}$ denotes the virtual flow between nodes $i$ and $j$, and $y_{ij}$ is a binary indicator for edge activation. 
The constraint violation term is defined via $P_{\text{connectivity}} = 1$  if  $\mathcal{S}$ is disconnected, and $0$ otherwise. 
A binary formulation is used to impose strict satisfaction of the connectivity requirement without allowing partial feasibility.

\paragraph{Coverage}
This constraint is to promote $r$-hop coverage of nodes \cite{10664002}, with critical nodes prioritized. It aims to encourage critical nodes visibility to the detection model, either directly equipped with a sensor or sufficiently covered by neighbors within $r$-hop reach. Each node \( j \) is said to be covered if there exists at least one sensor within its $r$-hop neighborhood:
\begin{equation}
\text{coverage}(j) = 
\begin{cases}
1, & \text{if } \sum_{i \in \mathcal{N}_r(j)} x_i \geq 1 \\
0, & \text{otherwise}
\end{cases}
\end{equation}
where \( \mathcal{N}_r(j) = \{ i \in \mathcal{V} \mid d(i, j) \leq r \} \) is the $r$-hop neighborhood of node $j$, i.e., the set of nodes that can cover node \( j \) within a coverage radius $r$, and $d(i,j)$ is the topological distance  between nodes $i$ and $j$, computed using the Breadth-First Search (BFS) algorithm. The corresponding penalty is computed as:
\(
P_{\text{coverage}} = \frac{1}{N_c} \sum_{j=1}^{N_c} \alpha_j \left(1 - \text{coverage}(j)\right),
\) 
where $\alpha_j$ is the criticality weight of node $j$.

\paragraph{Redundancy}
A single-sensor coverage may be insufficient to guarantee robustness against failures. Hence, a redundancy constraint is used to encourage nodes being monitored by multiple sensors \cite{10664002}.  Introducing \(u_j = \sum_{i \in \mathcal{N}_r(j)} x_i\),  the violation is expressed as: 
 \(
 P_{\text{redundancy}} = \frac{1}{N_c}  \sum_{j=1}^{N_c} \mathrm{max}\left(\mathrm R_{\text{min}} - u_j,0\right)
 \),
 where $R_{\text{min}}$ is the minimum redundancy requirement.

\section{Proposed Joint Optimization Framework}
\label{proposed_work}
\subsection{Optimization Methodology Using NSGA-II}
\label{method_opt}
The sensor placement strategy follows a principled approach that accounts for the criticality of nodes within the power network.   
We propose a hybrid population initialization strategy tailored for this placement optimization problem (Algorithm \ref{alg:pop_init}). It aims to generate a diverse and high-quality set of initial candidate solutions, each representing a valid configuration of sensors placed on a subset of network buses. It combines domain-driven heuristics with diversity-aware random sampling that improves search efficiency.
The initialization strategy starts with heuristic seeding in which a fraction of the population is initialized using top-\( K \) selection based on importance scores  and greedy coverage to maximize observability within a predefined radius. Then additional individuals are generated randomly, but only accepted if they meet a minimum Hamming distance threshold from existing solutions. Remaining individuals are filled using random sampling to reach the desired population size.

\begin{algorithm}[!t]
\footnotesize
\caption{Hybrid population initialization}
\label{alg:pop_init}
\begin{algorithmic}[1]
\Require graph \( \mathcal{G} \), eligible nodes \( \mathcal{V}_{\text{eligible}} \), importance scores \( S_{I} \), population size \( N_{\rm pop} \), maximum number of sensors \( K \), coverage radius \( r \), heuristic fraction \( h_{\text{frac}} \), diversity fraction \( d_{\text{frac}} \), Hamming threshold \( d_{\min}. \)

\Ensure  Initialize empty population \( \mathcal{P} \leftarrow \emptyset \), \( N_c \leftarrow |\mathcal{V}_{\text{eligible}}|. \)
\State Compute descending order of importance scores \( S_{I} \rightarrow {S_{I}}_{\text{sorted}} \)
\ForAll{\( i \gets 1 \) to \( \lfloor h_{\text{frac}} \cdot  N_{\rm pop} / 2 \rfloor \)}
    \State Select top-\( K \) nodes from  \(S_{I_{\text{sorted}}}\), apply tie-breaking if needed
    \State Add to \( \mathcal{P} \) if unique
\EndFor
\ForAll{\( i \gets 1 \) to \( \lceil h_{\text{frac}} \cdot  N_{\rm pop} / 2 \rceil \)}
    \State Apply greedy \( r \)-hop coverage from \( \mathcal{V}_{\text{eligible}} \)
    \State Add to \( \mathcal{P} \) if unique
\EndFor
\While{ \( |\mathcal{P}| < (h_{\text{frac}} + d_{\text{frac}}) \cdot  N_{\rm pop} \) }
    \State Randomly select \( \leq K \) nodes from \( \mathcal{V}_{\text{eligible}} \)
    \State Construct binary vector \( \mathbf{X} \)
    \If{Hamming(\( \mathbf{X}, \mathcal{P} \)) \( \geq d_{\min} \cdot N_c \)}
 \State Add \( \mathbf{X} \) to \( \mathcal{P} \)
    \EndIf
\EndWhile
\While{ \( |\mathcal{P}| < N_{\rm pop} \)}
    \State Randomly select \( \leq K \) nodes from \( \mathcal{V}_{\text{eligible}} \)
    \State Add to \( \mathcal{P} \) if unique
\EndWhile
\State \Return Population \( \mathcal{P} \) of binary vectors
\end{algorithmic}
\end{algorithm}
\begin{algorithm}[!t]
\footnotesize
\caption{{Biased mutation}}
\label{alg:biased_mutation}
\begin{algorithmic}[1]
\Require binary vector $\mathbf{X}$,
  importance scores $S_I$, 
  mutation probability ${indpb}$,  
  probability of importance-based flip $b_F$.
\Ensure  \textit{FlipBit}(.) be a random flip function,  and \textit{UniRand}(0, 1) be a uniformly-distributed random number generator function.
\ForAll{$i \gets 1$ to \text{length}($\mathbf{X}$) } \Comment{Nested mutation prob. logic}
    \If{$\text{\textit{UniRand}} < {indpb}$} \Comment{1. Main Mutation prob. logic}
  \If{$\text{\textit{UniRand}} < b_F$} \Comment{2. Importance-based logic}      
      \If{ $\text{\textit{UniRand}} < I[i]$} \Comment{3. Importance-based flip}
\State  $\mathbf{X}[i] \gets 1$ 
      \Else
\State $\mathbf{X}[i] \gets 0$
      \EndIf
  \Else
      \State $\mathbf{X}[i] \gets$ $\text{\textit{FlipBit}}(\mathbf{X}[i])$ \Comment{Flip randomly otherwise}
  \EndIf
  
 \EndIf
      
\EndFor
\State \Return $\text{Mutated } \mathbf{X}$
\end{algorithmic}
\end{algorithm}

Mutation operations within NSGA-II are guided by node importance (Algorithm \ref{alg:biased_mutation}), with the probability checks applied sequentially to implement independent random draws at each stage, and preserve stochastic diversity in the search process. Non-dominated sorting and crowding distance measures guide selection towards diverse, high-quality solutions. Offspring replace parent solutions based on Pareto dominance and solution diversity metrics, iterating until convergence. 
{NSGA‑II sorts the joint parent–offspring population into Pareto fronts based on dominance; within each front,  individuals are ranked by crowding distance, thus preserving a diverse approximation to the Pareto optimal set without requiring any user‑defined objective weights.}  The constraints are handled via Deb’s constraint‐domination principle \cite{deb2000efficient}. In NSGA-II each individual is evaluated on three objectives: $\left(V, \, f_1, \, f_2 \right)$.

During non‐dominated sorting, NSGA-II treats $V$ as the \emph{first} objective to minimize such that fully feasible solutions dominate infeasible ones and among infeasible solutions, dominance is determined by the degree of constraint violation. At termination, the Pareto front may contain both feasible (\(V_i=0\)) and infeasible (\(V_i>0\)) trade‐offs.  We select our final champion as follows: 
\(
\text{best} 
= \min_{i\in\text{front}} \left({\mathbb{I}(V_i>0)},\;
V_i,\;f_{1,i}+f_{2,i}\right) \). 
The selection process goes as follows: (i) all feasible solutions rank ahead of any infeasible, (ii) among infeasible, the one with smallest \(V_i\) is preferred, and (iii)  within the chosen class (all with equal first two values),  the individual with minimal \(\,f_{1,i}+f_{2,i}\) is selected. In this setup, \(\varepsilon\) is not explicitly predefined; its effective value emerges from the final Pareto front, being either zero when feasible solutions exist or equal to the smallest \(V_i\) among the infeasible solutions selected according to the defined ranking process.

\subsection{Node Importance Metric}

To identify critical buses in the power system network, we utilize a hybrid node importance score that combines both \textit{topological} and \textit{electrical} centrality metrics. 
For each bus \( v \in \mathcal{V} \), the importance score \( S_{I}(v) \) is computed as a weighted sum of individually normalized centrality values:
\(
S_{I}(v) =  w_{\text{bc}} \hat{C}_{\text{B}}(v) + w_{\text{eic}} \hat{C}_{\text{EI}}(v) + w_{\text{ebc}} \hat{C}_{\text{EBC}}(v) + w_{\text{ecd}}  \hat{C}_{\text{ECD}}(v),
\)
where \( \hat{C}_i(v) = C_i(v) / \max(C_i) \) is the normalized centrality score for metric \( i \), and \( w_i \) is the corresponding weight, chosen empirically based on ablation study results, satisfying \( \sum w_i = 1 \). These metrics are defined as follows: 

\subsubsection{Topological Centrality Metrics}
These metrics capture the structural role of buses \cite{ma2021deep}, and quantify node criticality based on graph connectivity. We employ: 
(i) {Betweenness centrality} which measures the extent to which a node lies on paths between other nodes. For a node \( v \), it is given by: \(      C_{\text{B}}(v) = \sum_{s \neq v \neq t} \frac{\sigma_{st}(v)}{\sigma_{st}},\) 
where \( \sigma_{st} \) denotes the total number of shortest paths from node \( s \) to node \( t \), and \( \sigma_{st}(v) \) stands for the number of those paths that pass through node \( v \).   {In this context, it highlights buses that act as structural bridges or bottlenecks.} (ii) {Eigenvector centrality} that measures node’s influence within a network, where connections to more central nodes contribute more to the score.  {It identifies buses with widespread topological influence, even if they do not have many direct connections.}
It is expressed as: \(C_{\text{EI}}(v) = \frac{1}{\lambda} \sum_{k \in \mathcal{N}(v)}  C_{\text{EI}}(k), \)     where \( \lambda \) is the largest eigenvalue of the adjacency matrix, and \( \mathcal{N}(v) \) is the set of neighbors of node \( v \).

\subsubsection{Electrical Centrality Metrics}
These metrics capture the physical and operational roles of buses in the power system \cite{7931591, wang2021method,bai2015hybrid}. We utilize two complementary metrics, for power transfer roles and for electrical coupling and proximity: 
(i) {Electrical betweenness centrality},  which accounts for actual power flow by quantifying the participation of a node in power transfers between generator--load pairs. It helps to identify buses that serve as key power conduits across the network. The following notations will be used: \( G \) for the set of generator nodes,  \( L \) for the set of load nodes,   \( W_i \) for the rated generation capacity of generator \( i \),    \( W_j \) for the actual or peak load at load node \( j \), 
\(
B_e(n) = \sum_{i \in G, j \in L} \sqrt{W_i W_j} B_{e,ij}(n),
\)
where  \( B_{e,ij}(n) \) is the electrical betweenness of node \( n \) under unit injection between node pair \( (i, j) \), defined as, 
\[
B_{e,ij}(n) = 
\begin{cases}
\frac{1}{2} \sum_m |I_{ij}(m, n)|, & n \neq i, j \\
1, & n = i \text{ or } j,
\end{cases}
\]  
\( m \) is a node directly connected to \( n \), and \( I_{ij}(m, n) \) is the current in link \( (m, n) \) under unit injection between \( i \) and \( j \). The normalized electrical betweenness centrality is expressed as:
\(
C_{\text{EBC}}(v) = \frac{B_e(v)}{\sum_{i \in G, j \in L} \sqrt{W_i W_j}}
\), and 
(ii) {Electrical coupling connection degree}, which represents electrical closeness using the resistance distance derived from the impedance matrix \( Z_{\mathrm{bus}} = Y_{\mathrm{bus}}^{-1} \). It measures how easily a node can reach or influence others in the network, i.e., electrically well-integrated buses that can quickly affect or respond to system changes. It is expressed as:
\(
C_{\text{ECD}}(v) = \frac{1}{\sum_{u \neq v} (Z_{vv} + Z_{uu} - 2Z_{vu})}. \)

\subsection{Proposed Detectors: Physics-Informed GNN-based Models}

\label{method_ml}

Given $N$ buses for different power network topologies, the following measurements are acquired for each bus $i$ 
\begin{equation}
    \mathcal{X}_i=[V_i,I_i,\theta_i,\delta_i,P_i,Q_i],
    \label{eq:x_i}
\end{equation}
where $V_i$ and $I_i$ denote voltage and current magnitudes, $\theta_i$ and $\delta_i$ stand for voltage and current phase angles, and $P_i$ and $Q_i$ represent active and reactive power measurements, respectively. 
In our forward modelling case, PINN is leveraged as a conventional numerical solver approach. Therefore, we assume that the unknown $u(\mathcal{X},t)$ is obtained by an \textit{ansatz} $u(\mathcal{X},t; \bm{w})$, consisting of the proposed GNN-based network with input features $\mathcal{X}= (x_1,...,x_d)$ in the domain $\mathcal{X}\in\Omega\subset R^d$, time $t$ incorporates the temporal correlations within the domain $[0,T]$ and weights $\bm{w}$ of the graph-based network. The open set $\Omega\subset R^d$ represents the physical domain, and $u$: $\Omega\cup\partial\Omega \rightarrow R$ is a physical quantity, with $\partial\Omega$ as the boundary of $\Omega$. The objective  is to obtain network weights $\bm{w}$ such that \cite{MCCLENNY2023111722}
\begin{equation}
    \begin{split}
    &\mathcal{F}(u(\mathcal{X},t;\bm{w}),\mathcal{X},t,Du(\mathcal{X},t;\bm{w}),D^2u(\mathcal{X},t;\bm{w}),...,\\ &D^ku(\mathcal{X},t;\bm{w});\phi)=f(\mathcal{X},t).\hspace{0.15in} \mathcal{X},t\in \Omega.
    \label{eq:Def1}
    \end{split}
\end{equation}
Equation (\ref{eq:Def1}) represents a general partial differential equation (PDE)  of order $k$, where $\mathcal{F}$ is the given function for the problem at hand, $f(\mathcal{X},t)$ is the forcing function, and the equation must be satisfed at all $\mathcal{X},t \in \Omega$. The partial derivatives of the GNN-based network output for $\mathcal{X}$ and $\bm{w}$ at time $t$ in  $\mathcal{F}$ {are calculated using automatic-differentiation ($\mathrm{AutoDiff}$) methods}.  They are described collectively by $D^\alpha u(\mathcal{X},t; \bm{w})$, which is the set of all the partial derivatives of $u$ of order $\alpha$:
\begin{equation}
    D^\alpha u(\mathcal{X},t;\bm{w}) = 
    \left\{ \frac{\partial^\alpha u}{\partial x_1^{\alpha_1} \cdots \partial x_d^{\alpha_d}} 
    \;\middle|\; \alpha_1 + \cdots + \alpha_d = \alpha \right\}.
\end{equation}
{$\mathrm{AutoDiff}$ is a computational method that allows  to obtain the exact derivatives in an efficient manner.}

In this work, an additional loss term is added to the GNN variants' loss function to constrain their outputs to satisfy the physics-driven laws (Kirchhoff's laws). For the data obtained from each bus (vector $\mathcal{X}$ in Equation (\ref{eq:x_i})), the outputs should be in line with Kirchhoff’s laws at each node and at each time slot $t$. This approach offers the advantage of identifying whether an attack is present or not along with the different types of attacks present at a given network topology, satisfying the physics-based relationships between measurements for the power flow calculations. Therefore, the physics-driven loss functions for the measurements at bus $i$ are defined as
\begin{equation}
    \mathcal{L}_P(\hat{\mathcal{X}})=\frac{1}{N}\sum_{k=1}^N\left|\hat{P}_i^k-\hat{V}_i^k\hat{I}_i^k\cos{\left(\hat{\theta}_i^k-\hat{\delta}_i^k\right)}\right|^2
\end{equation}
\begin{equation}
    \mathcal{L}_Q(\hat{\mathcal{X}})=\frac{1}{N}\sum_{k=1}^N\left|\hat{Q}_i^k-\hat{V}_i^k\hat{I}_i^k\cos{\left(\hat{\theta}_i^k-\hat{\delta}_i^k\right)}\right|^2,
\end{equation}
where $\mathcal{L}_P(\hat{\mathcal{X}})$ and $\mathcal{L}_Q(\hat{\mathcal{X}})$ are  the deviations of the outputs from $P$ and $Q$ in the AC power flow equations, respectively. The total function for training the PIGNN models is created by combining the data and physics loss functions as
\begin{equation}
\mathcal{L}_T(W,b,\mathcal{X},\hat{\mathcal{X}})=\lambda_{\mathrm{Data}}\mathcal{L}_{\mathrm{Data}}+\lambda_{\mathrm{Phy}}(\mathcal{L}_P+\mathcal{L}_Q),
\end{equation}
where $\lambda_{\mathrm{Data}}$ and $\lambda_{\mathrm{Phy}}$ denote the data-driven and physics-based loss coefficients. Such regularization  ensures the balance between the loss terms. {Through hyperparameter searching,  $\lambda_{\mathrm{Data}}$ was set to 1 and $\lambda_{\mathrm{Phy}}$ to 0.2.}

\section{Experimental Evaluation}
\label{res-discuss}
The proposed framework is evaluated on a power system already observable under its existing metering infrastructure that it is intended to supplement. {In baseline sensor layouts, \textit{generator buses} are equipped with voltage magnitude $\lvert V\rvert$ and active‐power injection $P$ measurements, (ii) \textit{load buses} provide active‐ and reactive‐power injections $(P,Q)$ readings, obtained either from dedicated meters or high‐quality pseudo‐measurements, and (iii) \textit{the slack (reference) bus} has voltage phasor $\lvert V\rvert\angle\delta$ together with active‐power injection $P$ \cite{glover2011power}. The data used in the development and evaluation phase  covers attack scenarios varying in the percentage of targeted buses. The training dataset contains random and general attack events, and a 70\% vs. 30\% split is used for the training and validation sets. A dedicated test dataset is collected for various unseen LR attack scenarios.   
The detection performance is evaluated using the following metrics, accuracy (ACC), true positive rate (TPR), false positive rate (FPR), precision (PREC), and F1-score. }
{The settings used for the optimization methodology are as follows: $N_{\rm pop}$ of 20--50, $b_F$ of 0.8, $indpd$ of 0.1,  $K$ as 30\% of number of buses, $h_{\rm frac}$ of 0.2, $d_{\rm frac}$ of 0.3, and $d_{\rm min}$ of 0.1--0.3. The constraints violations are given equal weights of 1, and the criticality node weights $\alpha$ are taken as the node importance scores.  } Through ablation analysis, the node importance score weights were determined to be set equally to 0.25, and the coverage radius $r$, and minimum redundancy $R_{\rm min}$  to 1 and 2, respectively.

\subsection{Detection Performance Against ML Baselines}
Table \ref{ML_benchmarking} presents the test performance of various models across the test cases. The evaluated models are graph-based models, including  GCNs, graph attention networks (GATs), Graph Sample and Aggregate (GraphSAGE), GTNs, graph diffusion networks (GDNs), PIGCNs, PIGDNs, and the proposed PIGTN. 
Physics-informed variants (PIGCN, PIGDN, and PIGTN) ensured that every training sample satisfied the AC-power flow equations (Kirchhoff's laws). Therefore, the additional physics-based loss term added to the original loss function constrained the output to satisfy the physics-driven laws. In the event of an attack, the subsets of measurements attributed to any type of the attacks investigated, imposed large AC-residuals. Consequently, these models retained the near-zero FPR of the base graph network, and recovered much higher TPR, pushing F1-score and overall accuracy to the top across all test cases. Embedding Kirchhoff’s laws acted as an inductive prior that suppressed the large AC residual patterns and allowed the network to generalize across different grid sizes and attack types. It substantiated the reason as to why the PIGCN and PIGTN were performing the best relative to the plain graph-based models. In fact, averaged across all seven tested power network topologies, PIGTN reached ACC $\approx$ 0.933, F1-score $\approx$ 0.928, with TPR $\approx$ 0.865, and FPR $\approx$ 0.0023, versus the best non-physics baseline here, GraphSAGE at ACC $\approx$ 0.9146, F1-score $\approx$ 0.9057, TPR $\approx$ 0.8326, and FPR $\approx$ 0.0033. More specifically, the performance improvements in the proposed PIGNN-based (physics $+$ topology-aware models) attack detectors compared to the other graph-based/topology-aware models (GraphSAGE, GAT, GTN, GCN, and GDN) for different bus system cases are listed as follows:
\begin{itemize}
    \item In the \textbf{14} bus system, PIGTN offered performance improvements of 2.3\%--17.4\% in ACC, 4.3\%--30.9\% in TPR, and 2.6\%--22.2\% in F1-score. 
    \item In the \textbf{30} bus system, PIGTN offered performance improvements of 1.1\%--14.3\% in ACC, 2.1\%--29.2\% in TPR, and 1.3\%--21.2\% in F1-score. 
    \item In the \textbf{IEEE 30} bus system, PIGTN offered performance improvements of 4.4\%--24.3\% in ACC, 8.6\%--49\% in TPR, and 5.2\%--37.9\% in F1-score. 
    \item In the \textbf{39} bus system, PIGTN offered performance improvements of 0.1\%--6.5\% in ACC, 0\%--12.8\% in TPR, and 0.1\%--7\% in F1-score. Notably, most models performed well, due to the moderate topological complexity and feature variability of this case. 
    \item In the \textbf{57} bus system, PIGTN offered performance improvements of 3.7\%--12.4\% in A.CC, 7.2\%--24.7\% in TPR, and 4.3\%--16.5\% in F1-score. 
    \item In the \textbf{118} bus system, PIGTN offered performance improvements of 0.4\%--36.4\% in ACC, 0.9\%--73\% in TPR, and 0.6\%--73.2\% in F1-score. 
    \item In the \textbf{200} bus system, PIGTN offered performance improvements of 1.1\%--26.5\% in ACC, 1.2\%--52\% in TPR, and 1.8\%--44.5\% in F1-score. 
\end{itemize}

The added AC power flow loss constraints, forces hidden states to abide by Kirchhoff's laws by penalizing mismatches in $P$ and $Q$. Attacks that locally tamper with the measurements while leaving global power-flow relations inconsistent produce larger AC residuals, where the model learns such correlations as these inconsistencies are separable from normal variability.

\begin{table}
\centering
\scriptsize
\caption{Test-set performance of models across different bus system cases.}
\label{ML_benchmarking}
 \resizebox{\linewidth}{!}{%
\begin{tabular}{llccccccccccc}
\toprule
\textbf{Case}& \textbf{Metric} &  \textbf{GraphSAGE} & \textbf{GAT} &  \textbf{GCN} & \textbf{GDN} & \textbf{GTN} & \textbf{PIGCN}  & \textbf{PIGDN} & \textbf{PIGTN} \\
\midrule
\textbf{Case 14} & ACC  & \underline{0.927} & 0.776        & 0.871        & 0.883        & 0.829        & 0.922                          & 0.842          & \textbf{0.950} \\
& PREC &  0.994                          & 0.933        & 0.989        & 0.991        & 1.000        & 0.994                          & 0.990          & 0.997          \\
& TPR  &  0.860                          & 0.594        & 0.750        & 0.772        & 0.658        & 0.849                          & 0.690          & 0.903          \\
& FPR   &  0.0054                         & 0.042        & 0.008        & 0.007        & 0.000        & 0.005                          & 0.0069         & 0.0023         \\
& F1   &  0.922                          & 0.726        & 0.853        & 0.868        & 0.794        & 0.916                          & 0.814          & 0.948          \\ 
\midrule
\textbf{Case 30} & ACC  & \underline{0.886} & 0.761        & 0.851        & 0.876        & 0.754        & 0.873                          & 0.830          & \textbf{0.897} \\
&PREC & 0.989                          & 0.994        & 0.982        & 0.987        & 0.999        & 0.991                          & 0.999          & 0.991          \\
& TPR   & 0.780                          & 0.524        & 0.713        & 0.762        & 0.509        & 0.753                          & 0.661          & 0.801          \\
& FPR   & 0.0083                         & 0.0031       & 0.012        & 0.010        & 0.0004       & 0.007                          & 0.001          & 0.0075         \\
& F1   & 0.873                          & 0.687        & 0.827        & 0.860        & 0.674        & 0.855                          & 0.795          & 0.886          \\ 
\midrule
\textbf{Case IEEE 30} & ACC  &   0.886                          & 0.687        & 0.818        & 0.817        & 0.780        & \underline{0.910} & 0.857          & \textbf{0.930} \\
& PREC  &  0.991                          & 0.998        & 0.999        & 0.996        & 1.000        & 0.996                          & 0.987          & 0.994          \\
& TPR   &  0.779                          & 0.375        & 0.637        & 0.636        & 0.560        & 0.823                          & 0.723          & 0.865          \\
& FPR &  0.0068                         & 0.0008       & 0.0002       & 0.0025       & 0.000        & 0.0037                         & 0.0093         & 0.0052         \\
& F1  &  0.873                          & 0.546        & 0.778        & 0.817        & 0.718        & 0.901                          & 0.835          & 0.925          \\ 
\midrule
\textbf{Case 39} & ACC  & \underline{0.994} & 0.972        & 0.930        & 0.934        & 0.985        & 0.987                          & 0.985          & \textbf{0.995} \\
& PREC  & 0.999                          & 1.000        & 0.998        & 0.998        & 1.000        & 1.000                          & 1.000          & 1.000          \\
& TPR  & 0.989                          & 0.944        & 0.861        & 0.870        & 0.971        & 0.973                          & 0.969          & 0.989          \\
& FPR & 0.0002                         & 0.00         & 0.0017       & 0.0021       & 0.000        & 0.000                          & 0.000          & 0.000          \\
& F1 & 0.994                          & 0.971        & 0.925        & 0.929        & 0.985        & 0.986                          & 0.984          & 0.995          \\ 
\midrule
\textbf{Case 57} & ACC  & 0.894                          & 0.851        & 0.867        & 0.817        & 0.807        & \underline{0.911} & 0.859          & \textbf{0.931} \\
& PREC & 0.999                          & 0.983        & 0.998        & 0.999        & 0.999        & 1.000                          & 0.999          & 0.999          \\
& TPR   & 0.789                          & 0.714        & 0.736        & 0.635        & 0.614        & 0.822                          & 0.718          & 0.861          \\
& FPR  & 0.0004                         & 0.0123       & 0.0014       & 0.0004       & 0.0002       & 0.000                          & 0.0002         & 0.0002         \\
& F1    & 0.882                          & 0.827        & 0.847        & 0.776        & 0.760        & 0.902                          & 0.836          & 0.925          \\ 
\midrule
\textbf{Case 118} & ACC  &  \underline{0.907} & 0.811        & 0.794        & 0.892        & 0.547        & 0.873                          & 0.853          & \textbf{0.911} \\
& PREC &  0.999                          & 0.975        & 0.999        & 0.997        & 1.000        & 1.000                          & 0.999          & 0.999          \\
& TPR  &  0.814                          & 0.637        & 0.588        & 0.786        & 0.093        & 0.746                          & 0.706          & 0.823          \\
& FPR   & 0.001                          & 0.0164       & 0.0002       & 0.0021       & 0.000        & 0.000                          & 0.001          & 0.0008         \\
& F1   &  0.897                          & 0.771        & 0.740        & 0.879        & 0.171        & 0.855                          & 0.827          & 0.903          \\ 
\midrule
\textbf{Case 200} & ACC  & \underline{0.908} & 0.903        & 0.817        & 0.883        & 0.654        & 0.885                          & 0.845          & \textbf{0.919} \\
& PREC  & 0.999                          & 0.994        & 0.999        & 0.999        & 1.000        & 1.000  & 0.999          & 0.999          \\
& TPR & 0.817                          & 0.808        & 0.634        & 0.765        & 0.309        & 0.769                          & 0.691          & 0.829          \\
& FPR  & 0.001                          & 0.0031       & 0.0004       & 0.0004       & 0.000        & 0.000                          & 0.006          & 0.0003         \\
& F1    & 0.899                          & 0.892        & 0.776        & 0.867        & 0.472        & 0.870                          & 0.817          & 0.917          \\ 
\bottomrule
\end{tabular}%
 }
\end{table}

\begin{figure*}[ht]
    \centering
    
    \subfigure[Case 14]{\includegraphics[width=0.2\linewidth, trim=0 0 0 1, clip]{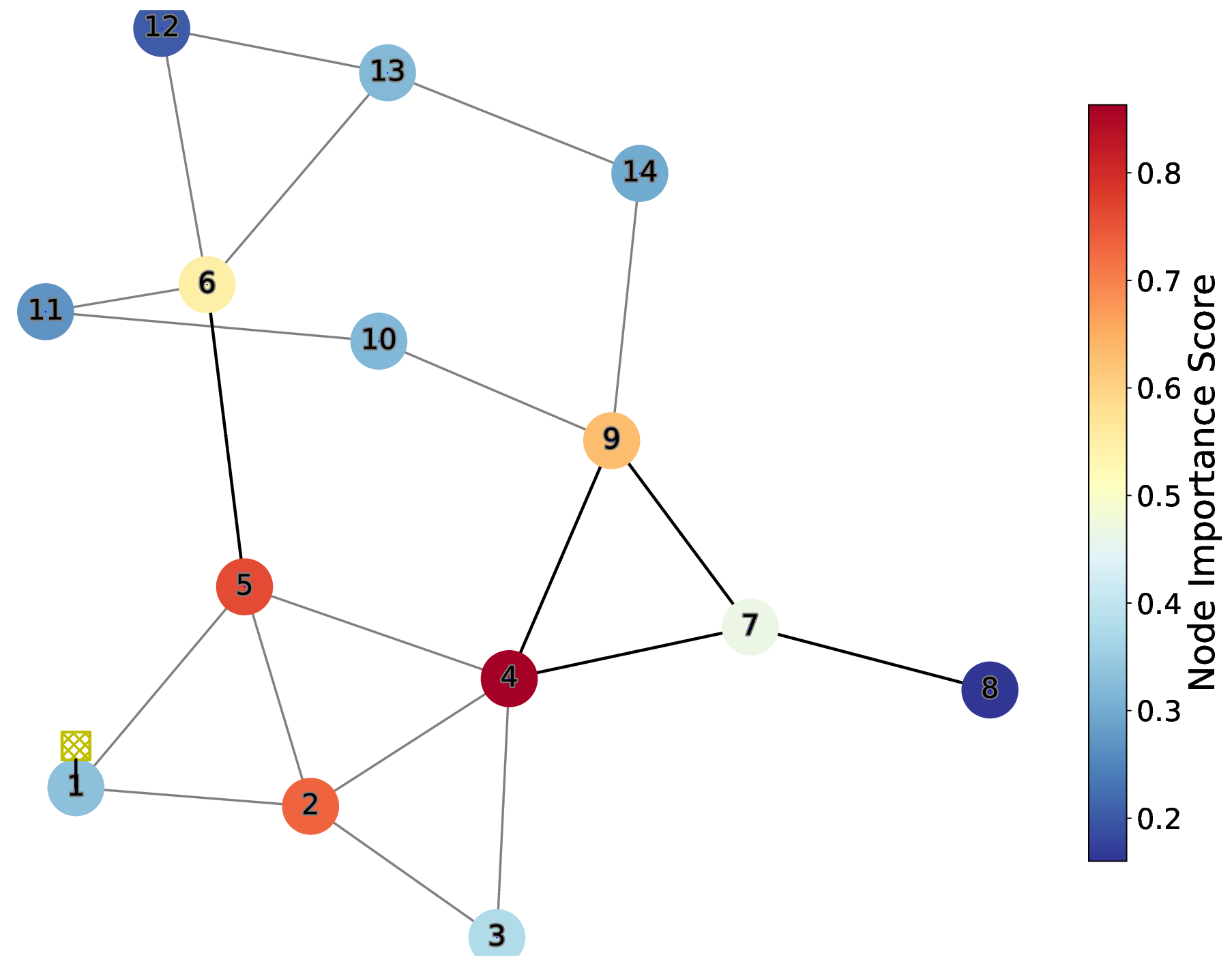}}
    \subfigure[Case 30]{\includegraphics[width=0.2\linewidth, trim=0 0 0 5, clip]{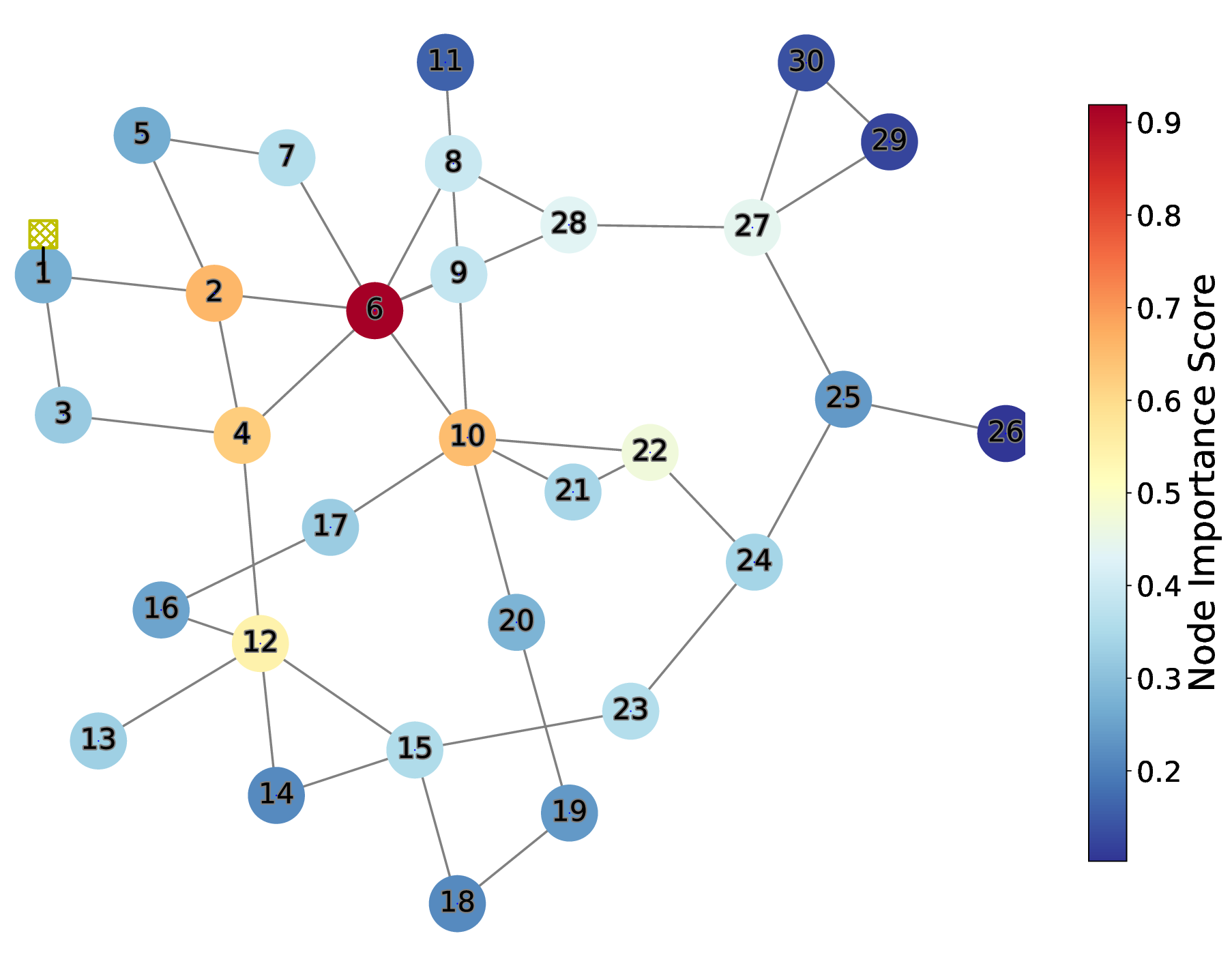}} 
    \subfigure[Case IEEE 30]{\includegraphics[width=0.2\linewidth, trim=0 0 0 5, clip]{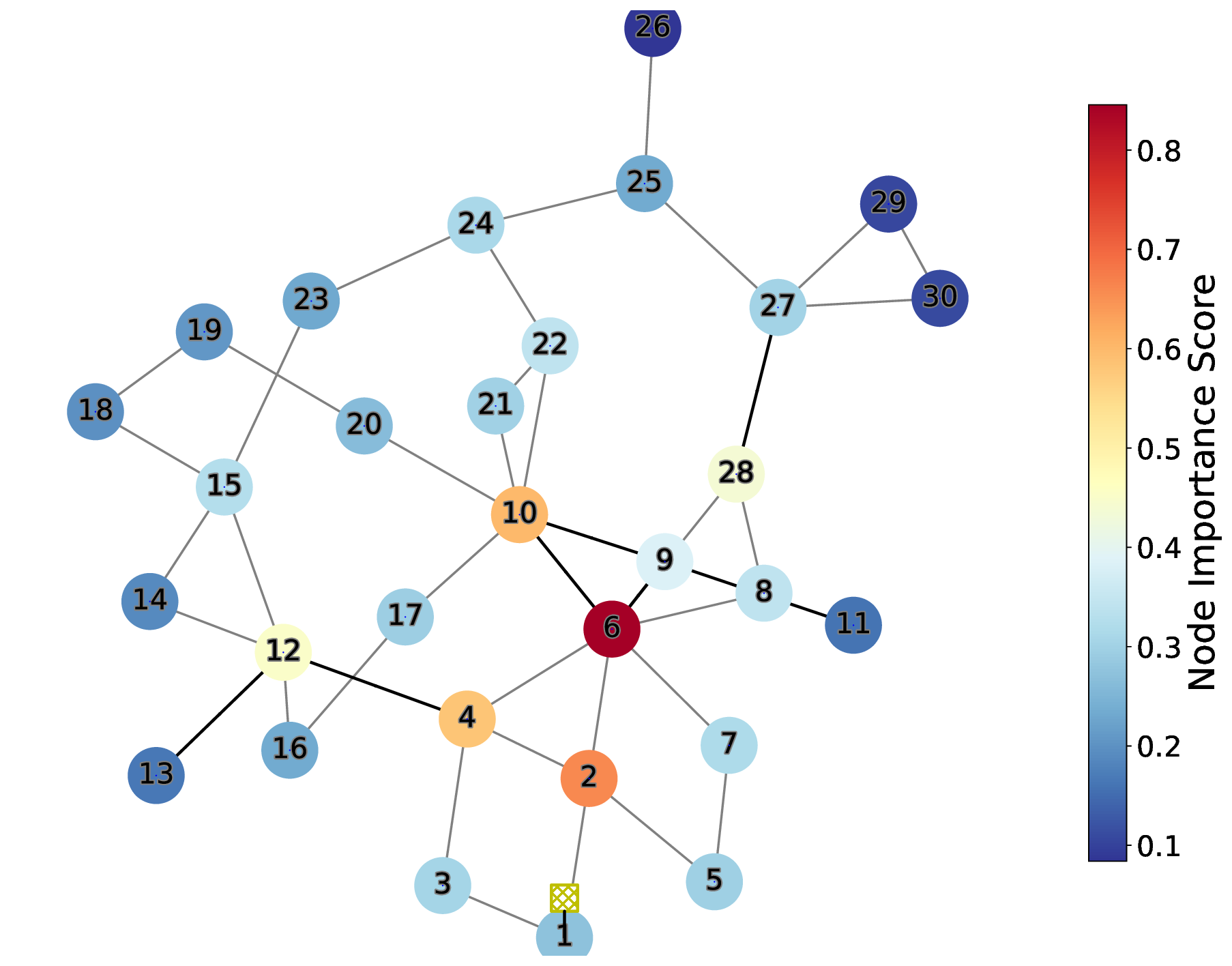}}
    \subfigure[Case 39]{\includegraphics[width=0.2\linewidth, trim=0 0 0 5, clip]{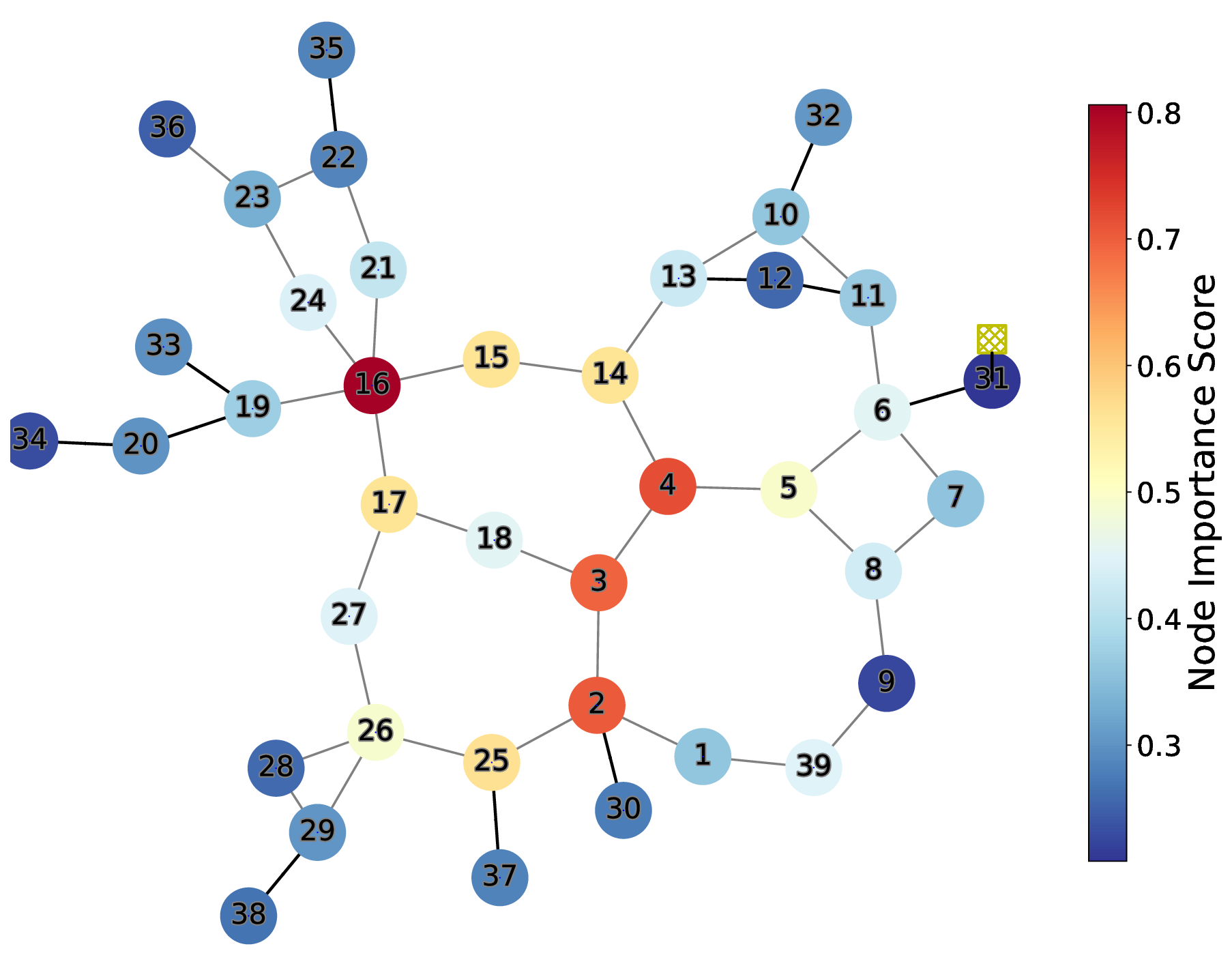}} \\
    \subfigure[Case 57]{\includegraphics[width=0.28\linewidth, trim=0 0 0 5, clip]{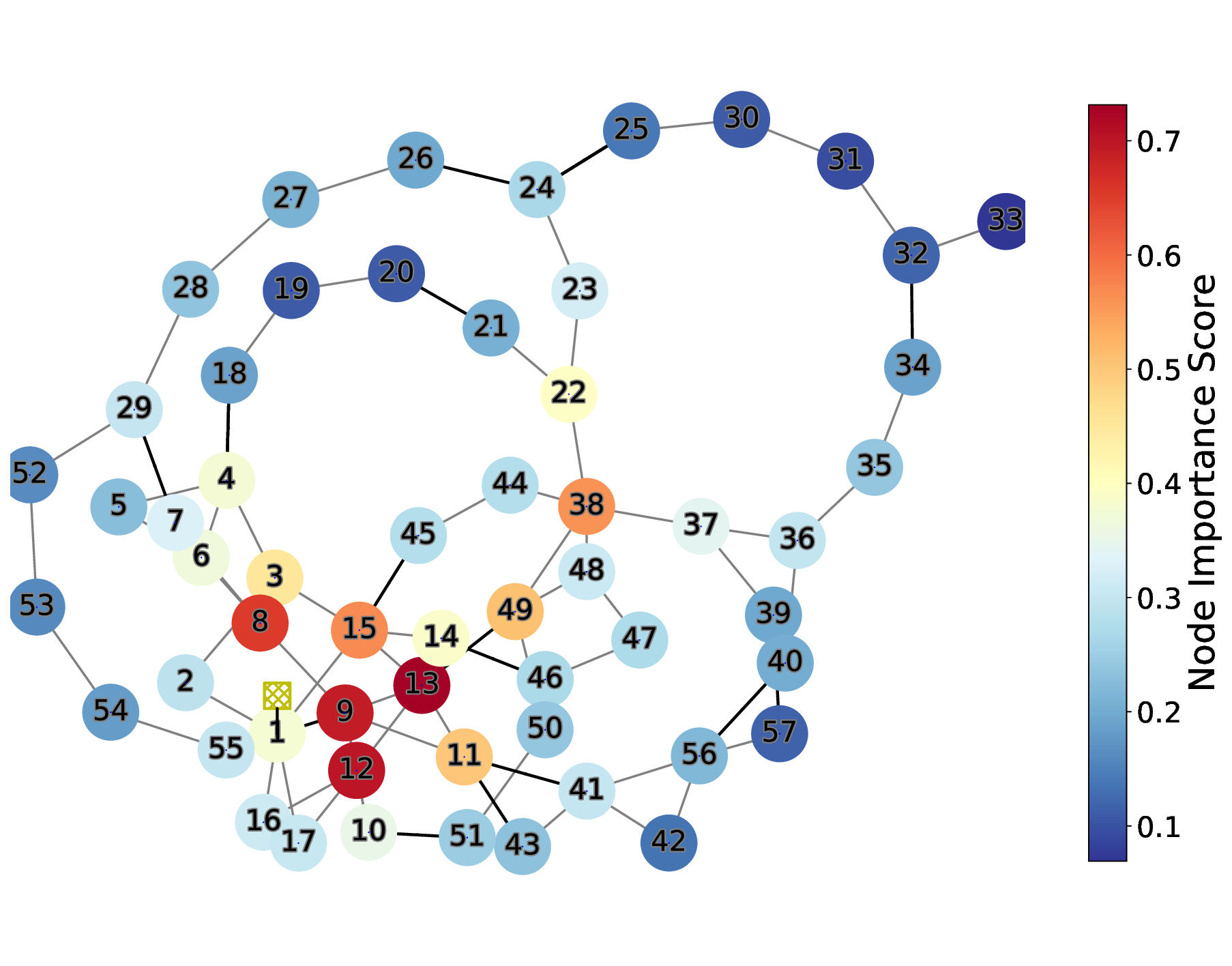}} 
    \subfigure[Case 118]{\includegraphics[width=0.28\linewidth, trim=0 0 0 5, clip]{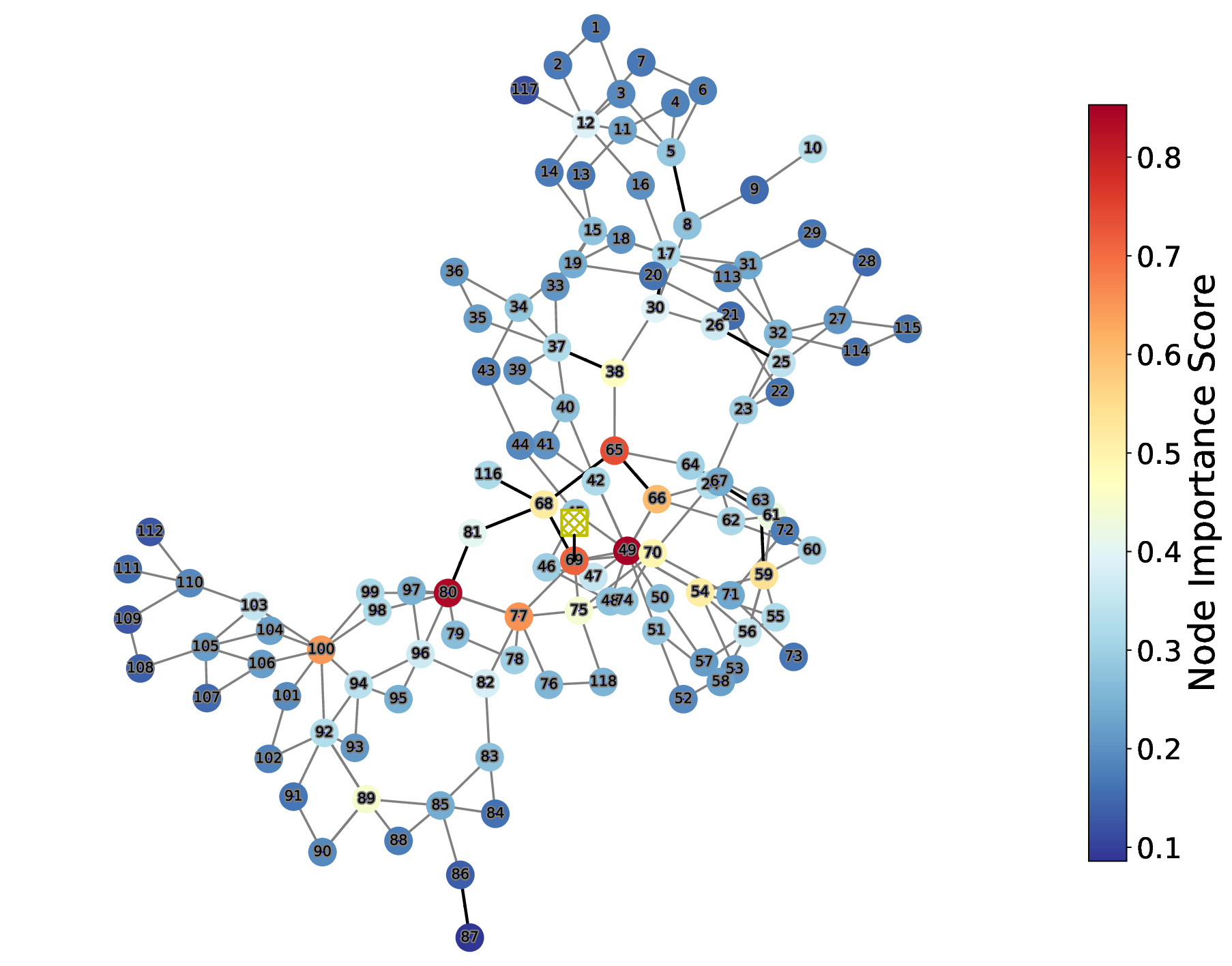}}
    \subfigure[Case 200]{\includegraphics[width=0.28\linewidth, trim=0 0 0 5, clip]{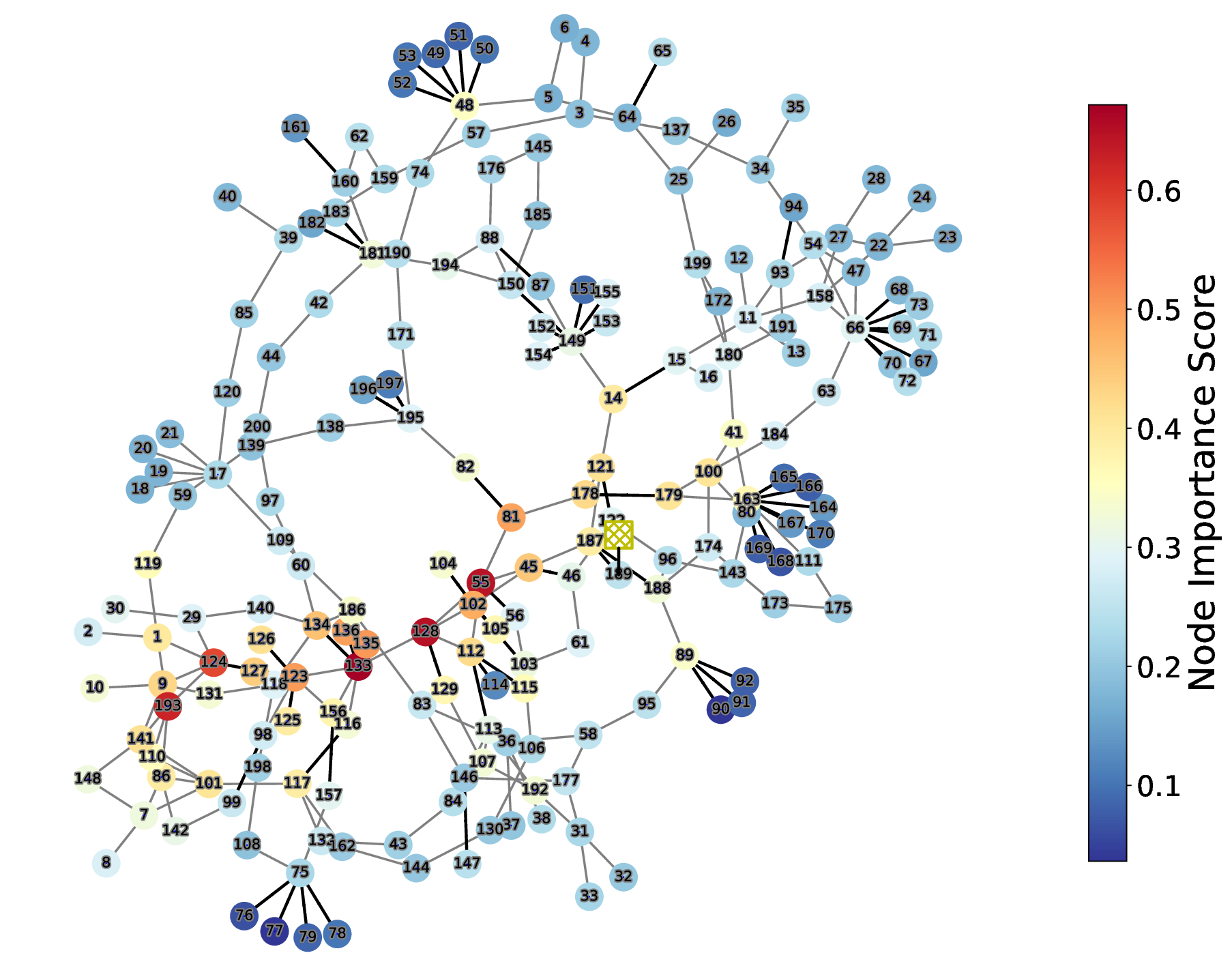}}
    
    \caption{Graph representations of cases (nodes colored according to computed importance scores).}
    \label{fig:topologies}
    
\end{figure*}

\begin{table*}[ht!]
    \centering
    \scriptsize
\caption{Sensor placements across test cases.} 
\label{sensor_placements_methods_res}
\resizebox{\linewidth}{!}{
\begin{tabular} {p{1.9cm}p{0.8cm}p{2.2cm}p{2.2cm}p{2.8cm}p{5cm}}
\toprule
 { \begin{tabular}{@{}c@{}}Placement method\end{tabular}} &  {Case 14} &  {Case 30} &  {Case IEEE 30} &  {Case 39} &  {Case 57}  \\
\hline
 {Ours} & 4,5,10,13 & 3,6,7,9,10,12,15,25,29 & 3,6,7,9,10,12,15,25,27 & 1,2,5,6,10,14,15,18,19,22,26 & 4,7,10,11,14,16,17,22,23,29,32,37,38,48,49,55,56  \\
 {Greedy} & 4,5,7,9 & 4,6--10,12,15,28 & 4,6,7,9,10,12,15,22,28 & 2--5,14--18,25,26 & 4,7,10,11,13--17,22,23,29,37,38,48,49,55  \\
 {Yildiz  \&  Abur \cite{10640266} }& 6 & 9,13,26,30 & 26,30 & 6,8,10,20,28,36 & 7,33  \\
\bottomrule
\end{tabular}}
\end{table*}
\begin{table*}[ht!]
    \centering
    \scriptsize
\begin{tabular} {p{1.9cm}p{6.2cm}p{9cm}}
\toprule
 { \begin{tabular}{@{}c@{}} Placement method\end{tabular}} &  {Case 118} & {Case 200} \\
\hline
 {Ours} & 2,5,9,11,17,20,23,28,35,37,45,47,48,50,51,57,58,60,63,64,67, 68,71,78,79,83,93--98,106,109,118 & 1,3,7,9,10,14,15,17,22,29,30,36,41,45,46,48,56,61,66,75,81,82,86,89,100--103,107,110,112,113,116--119,121--123,129,131,134,141,142,148,149,156,157,163,178,179--181,184,186--188,192,194,195 \\
 {Greedy} &  5,11,17,23,30,35,37,38,45,47,48,50,51,57,58,60,63,64,67--69,71,75,78,79,81--83,94--98,106,118 & 1,7--10,14,15,29,30,41,45,46,48,55,56,61,66,81,82,86,89,100-103,107,110,112,113,116-119,121-124,128,129,131,133,134,141,142,148,149,156,157,163,178--181,184,186-188,192--195 \\
 {Yildiz  \&  Abur \cite{10640266} }& 32,43,59,73,111,117 & 2,4,6,8,10,12,13,16,18,--21,23,24,26,28,30,32,33,35,37,38,40,88,162 \\
\bottomrule
\end{tabular}
\end{table*}

\begin{figure*}[ht!]
    \centering
    
    \subfigure[Case 14]{\includegraphics[width=0.24\linewidth, trim=0 0 0 20, clip]{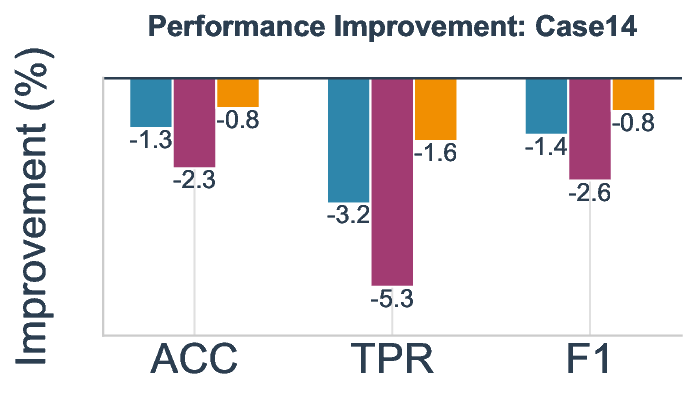}}
    \subfigure[Case 30]{\includegraphics[width=0.22\linewidth, trim=30 0 0 20, clip]{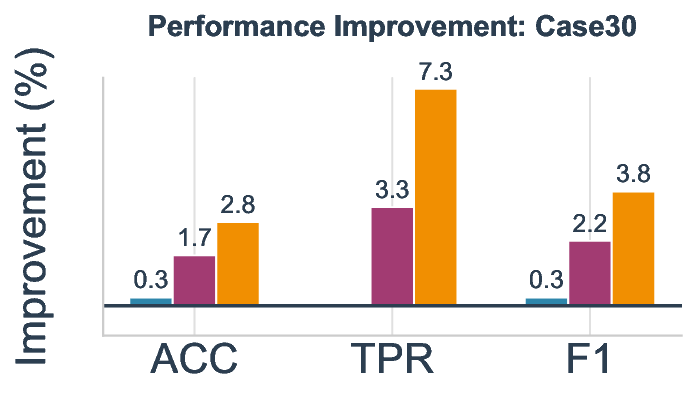}} 
    \subfigure[Case IEEE 30]{\includegraphics[width=0.22\linewidth, trim=30 0 0 20, clip]{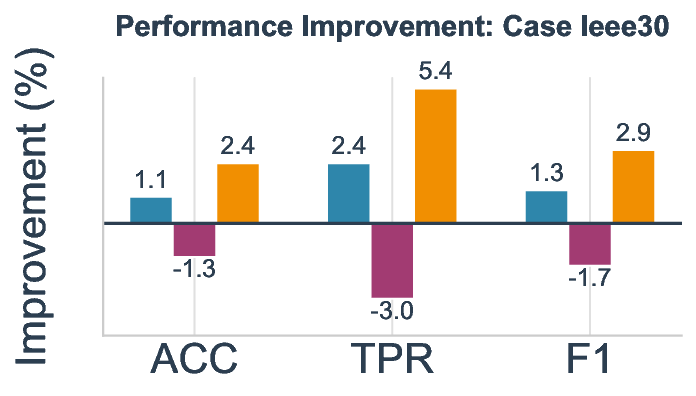}}
    \subfigure[Case 39]{\includegraphics[width=0.22\linewidth, trim=30 0 0 20, clip]{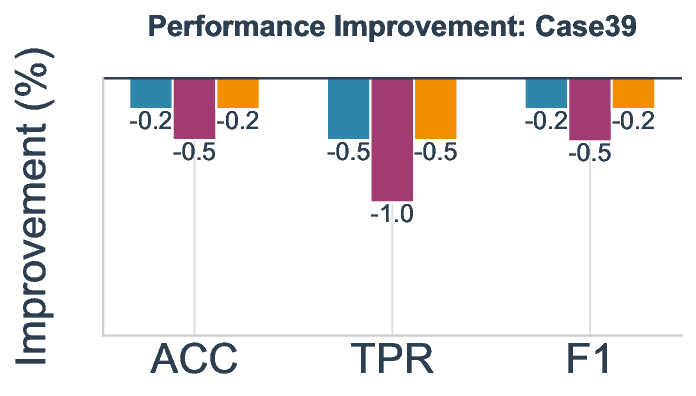}} \\
    \subfigure[Case 57]{\includegraphics[width=0.22\linewidth, trim=0 0 0 20, clip]{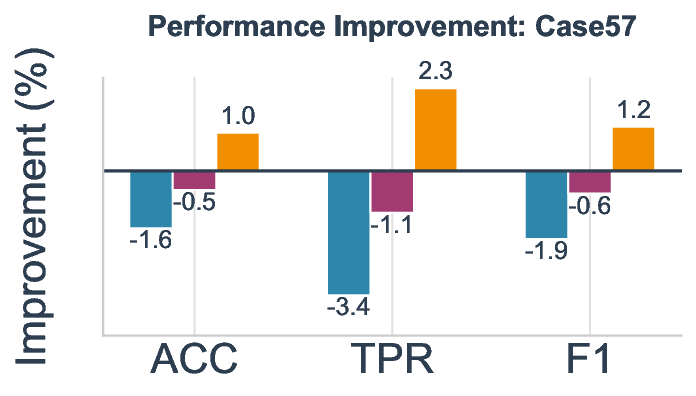}} 
    \subfigure[Case 118]{\includegraphics[width=0.22\linewidth, trim=30 0 0 20, clip]{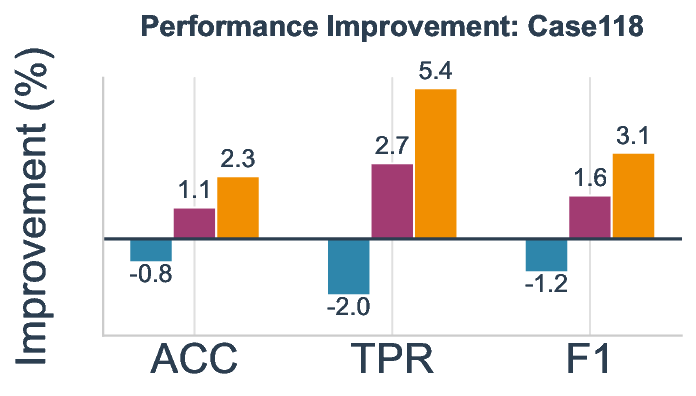}}
    \subfigure[Case 200]{\includegraphics[width=0.22\linewidth, trim=30 0 0 20, clip]{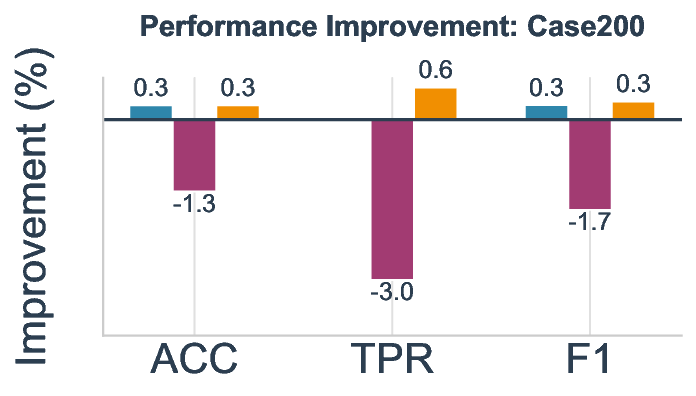}}
    \subfigure{\includegraphics[width=0.22\linewidth, trim=0 0 0 0, clip]{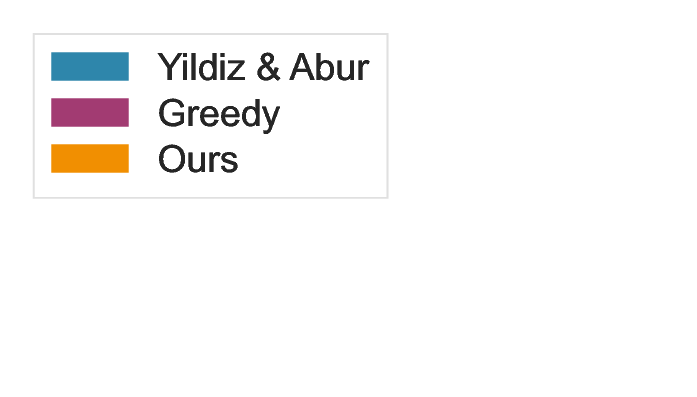}}
    
    \caption{Detection performance improvement across test cases compared to the baseline.}
    \label{fig:case_improvements}
\end{figure*}

\begin{table}[t]
\centering
\scriptsize
\caption{Evaluation on robustness under sensor failures across different bus system cases.}
\label{compare_with_placement_bn}
\resizebox{\linewidth}{!}{%
\begin{tabular}{llcccccc}
\toprule
\textbf{Case} & \textbf{Placement Method} & $F_{\text{crit}}$ & $\mathcal{R}$ & $\mathcal{A}_{\text{F1-score}}$ & $\overline{\mathrm{ACC}}$ & $\overline{\mathrm{F1}}$ & $\overline{\mathrm{PREC}}$ \\
\midrule
\multirow{4}{*}{\textbf{Case 14}}            & Ours   & 2 & \textbf{0.907} & \textbf{0.411}            & \textbf{0.799}            & \textbf{0.822}           & \textbf{0.743}\\
         & Greedy & 2 & \underline{0.900}    & \underline{0.401}  & \underline{0.776}  & \underline{0.803} & \underline{0.724}\\
         & Yildiz \& Abur            & 2 & 0.890          & 0.401& 0.766& 0.802    & 0.704 \\
         & Baseline  & 1 & 0.876          & 0.397& 0.749& 0.795    & 0.681 \\ 
         \midrule
\multirow{4}{*}{\textbf{Case 30}}      & Ours & 1 & \textbf{0.851} & \textbf{0.555}            & \textbf{0.672}            & \textbf{0.740}           & \textbf{0.624}\\
         & Greedy    & 1 & \underline{0.841}    & \underline{0.550}  & \underline{0.673}  & \underline{0.737} & \underline{0.629}\\
         & Yildiz \& Abur            & 1 & 0.841          & 0.547& 0.666& 0.731    & 0.625 \\
         & Baseline  & 1 & 0.826          & 0.535& 0.632& 0.716    & 0.591 \\ 
         \midrule
\multirow{4}{*}{\textbf{Case IEEE 30}} & Ours & 2 & \textbf{0.888} & \textbf{0.585}            & \textbf{0.747}            & \textbf{0.782}           & \textbf{0.704}\\
         & Greedy    & 2 & 0.840          & 0.568& 0.701& 0.759    & 0.647 \\
         & Yildiz \& Abur            & 2 & 0.864          & 0.581& 0.730& 0.775    & 0.681 \\
         & Baseline  & 2 & \underline{0.867}    & \underline{0.578}  & \underline{0.726}  & \underline{0.774} & \underline{0.685}\\ 
         \midrule  
\multirow{4}{*}{\textbf{Case 39}}      & Ours & 3 & \textbf{0.841} & \textbf{0.652}            & \textbf{0.769}            & \textbf{0.817}           & \textbf{0.697}\\
         & Greedy    & 3 & 0.840          & 0.647& 0.766& 0.813    & 0.694 \\
         & Yildiz \& Abur            & 3 & \underline{0.840}    & \underline{0.651}  & \underline{0.767}  & \underline{0.816} & \underline{0.697}\\
         & Baseline  & 3 & 0.828          & 0.642& 0.749& 0.804    & 0.676 \\ 
         \midrule
\multirow{4}{*}{\textbf{Case 57}}      & Ours & 2 & \textbf{0.828} & \textbf{0.647}            & \textbf{0.694}            & \textbf{0.759}           & \textbf{0.647}\\
         & Greedy    & 2 & 0.815          & 0.651& 0.692& 0.760    & 0.641 \\
         & Yildiz \& Abur            & 2 & \underline{0.821}    & \underline{0.645}  & \underline{0.691}  & \underline{0.755} & \underline{0.642}\\
         & Baseline  & 2 & 0.803          & 0.636& 0.666& 0.746    & 0.620 \\ 
         \midrule
\multirow{4}{*}{\textbf{Case 118}}     & Ours & \textbf{13}       & \textbf{0.981} & \textbf{0.778}            & \textbf{0.836}            & \textbf{0.833}           & \textbf{0.870}\\
         & Greedy    & 12& 0.957          & 0.776& \underline{0.831}  & 0.831    & \underline{0.846}\\
         & Yildiz \& Abur            & 12& \underline{0.981}    & \underline{0.778}  & 0.829& \underline{0.833} & 0.841 \\
         & Baseline  & 8 & 0.922          & 0.744& 0.776& 0.797    & 0.762 \\ 
         \midrule
\multirow{4}{*}{\textbf{Case 200}}     & Ours & \textbf{16}       & \textbf{0.912} & \textbf{0.782}            & \textbf{0.800}            & \textbf{0.815}           & \textbf{0.787}\\
         & Greedy    & \underline{11}          & \underline{0.887}    & \underline{0.778}  & \underline{0.784}  & \underline{0.812} & \underline{0.750}\\
         & Yildiz \& Abur            & 8 & 0.866          & 0.758& 0.758& 0.791    & 0.720 \\
         & Baseline  & 9 & 0.880          & 0.766& 0.773& 0.799    & 0.743 \\ 
         
\bottomrule
\end{tabular}
}
\end{table}

\subsection{Benchmarking Sensor Placement Strategies }

The evaluation carried out presents the improvement gain over the baseline for the GA-PIGTN framework and the benchmark placement methods, namely:  (i) the greedy placement method, which selects the top buses based on their importance score, and   (ii) the sparse placement method by Yildiz \& Abur \cite{10640266}, which is designed to yield the best possible locations for installing a limited number of PMUs to optimize their utility for model-based sparse estimation methods, provided that the baseline represents the existing sensing infrastructure.   Table \ref{sensor_placements_methods_res} and Fig. \ref{fig:topologies} show that GA placements presented more distributed and diverse sensor layouts compared to benchmark methods. It explored a broader search space, which shall also explain the superior results in large and structurally diverse networks. Greedy methods tend to cluster sensors at important nodes, while the Yildiz \& Abur method is more relevant in model-based detection settings. From the standpoint of detection under nominal operating conditions, Fig. \ref{fig:case_improvements} shows that GA-PIGTN placement exhibits varying performance across the test cases: (i) in Case 30, IEEE 30, and 118, the gain is most prominent, with 2.3\%--2.8\% improvements in ACC, 5.4\%--7.3\% in TPR, and 2.9\%--3.8\% in F1-score, (ii) in Case 57, and 200 the improvements are marginal, and (iii) in Cases 14 and 39, slightly negative gains across ACC, TPR, and F1-score were observed.     

 These results can be explained by the intrinsic characteristics of the systems. Relatively smaller and less complex topologies  (e.g., Cases 14, 39, and 57) that have high percentage of load and/or generator buses, inherently provide extensive measurements of active and reactive power--inferred, either directly or through pseudo-measurements. In such scenarios, the incremental contribution of optimized placements is smaller, leading to the modest or mixed gains observed.   Case 200, despite being the largest network, also exhibits only marginal detection gains. With a much dense and uniform topology,  GA-PIGTN placement added mostly redundant information from the current detection standpoint.  On the other hand, Case 118, which is modular with several loosely coupled sub-areas and corridor buses demonstrated a different behavior. 
 It showed noticeable improvements in detection due to its topological structure and heterogeneity. For Cases 30 and IEEE 30, well-connected topology with $\approx $ 70\% load buses and $\approx$ 17\% generator buses, the detection improvement gains were the most prominent. These results emphasized that the benefits of GA-PIGTN placement depend on the structural and functional characteristics of the system that can make additional sensors impactful or redundant to the detection model. Importantly, the GA-PIGTN framework consistently outperformed Greedy and Yildiz \& Abur  placements by yielding the least negative, more positive, and most prominent gains relative to the baseline sensor layout. This demonstrates the advantage of the joint optimization framework. Additional sensor placements do not necessarily lead to improved performance as meaningful gains are contingent on an effective and purposeful sensor placement, besides accounting for the intrinsic characteristics of the topology as discussed earlier.
 
On the other hand and across all cases, Table \ref{compare_with_placement_bn} consistently demonstrates the benefits of GA-PIGTN placements in terms of detection robustness. Random sensor failures were simulated that can range up to 30\% of the total number of buses.  Across smaller and medium-scale systems, GA-PIGTN provided a clear robustness advantage. In Cases 14, 39, and 57, GA-PIGTN still achieved either the highest or tied-best performance across all performance metrics.  Similarly, in Cases 30 and IEEE 30,  the GA-PIGTN scored the highest robustness performance compared to benchmarks. As network scale and connectivity play a critical role, GA-PIGTN placements exploited graph redundancy more effectively in Cases 118 and 200 and resulted in sensor layouts that increased resilience under sensor failures.    This was demonstrated by the higher $F_{\text{crit}}$ values of 13 and 16 sensors for Cases 118 and 200, respectively. While the GA-PIGTN performance superiority in terms of $F_{\text{crit}}$ was only evident in larger cases, the analysis of the remaining robustness metrics presented the following results against the baseline:
\begin{itemize}
    \item In terms of robustness score ($\mathcal{R}$) expressed  as $\mathcal{R} = 1 / \left ( 1 + \overline{\Delta \mathrm{ACC}} \right )$, where $\overline{\Delta \mathrm{ACC}}$ is the mean relative accuracy degradation over the failure range, GA-PIGTN showed improved performance by 3.4\% as opposed to the greedy and the Yildiz\&Abur, which showed 1.3\% and 1.5\% improvements, respectively.
    \item In terms of the normalized area under the curve of the F1-score versus failure level ($\mathcal{A}_{\text{F1}}$), GA-PIGTN resulted in improvements by 2.6\% vs. 1.4\%--1.6\% by the placement benchmarks.
    \item In terms of the average accuracy computed across  sensor failure levels  ($\overline{\mathrm{ACC}}$), GA-PIGTN demonstrated the best performance with 4.8\% improvements, while both greedy and Yildiz\&Abur showed 3\% improvements.
    \item In terms of the average F1-score ($\overline{\mathrm{F1}}$), GA-PIGTN resulted in improvements by 2.6\% vs. 1.3\%--1.6\%  by the placement benchmarks.
    \item In terms of the average precision ($\overline{\mathrm{P}}$), GA-PIGTN resulted in improvements by 6.5\% vs. 3.2\%--3.7\% by the placement benchmarks.
\end{itemize}

Taken together, even when detection accuracy gains are not significant, GA-PIGTN placements strengthen the resilience of the detection framework under stress conditions. Specifically, (i) in heterogeneous topologies, optimized placements added discriminative coverage and boosted accuracy; (ii) in uniformly structured, highly observable networks, added sensors were mostly redundant, improving robustness more than accuracy, and (iii) with moderate or imbalanced observability, GA-PIGTN bridges coverage gaps most effectively, resulting in significant gains. 

\begin{figure}[htp]
    \centering 
    
    \includegraphics[scale=0.35, trim=0cm 0.2cm 0.cm 0.5cm, clip]{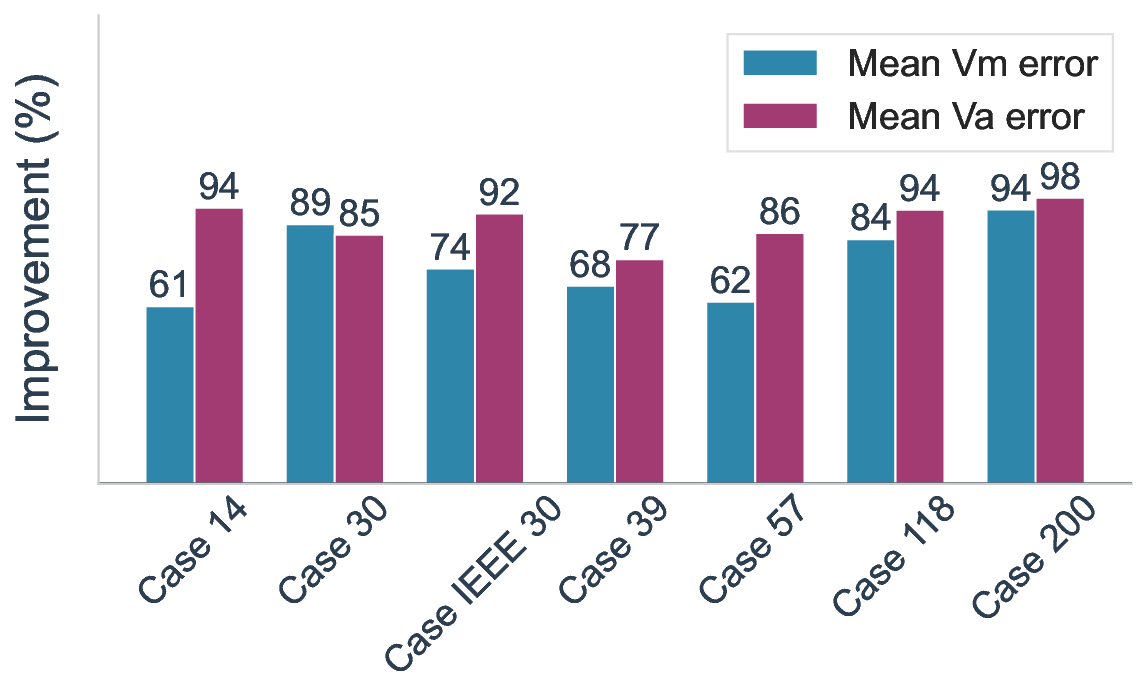}
    
    \caption{PSSE performance improvement across test cases.}
    \label{fig:SE_improve}
\end{figure}

\subsection{PSSE Performance Analysis}

In addition to the enhanced detection performance, the proposed framework led to improved state estimation performance in all test cases. Fig. \ref{fig:SE_improve} illustrates the improvement in PSSE performance, measured by the reduction in the mean voltage magnitude (Vm) and voltage angle (Va) errors. The improvement was evaluated on the WLS PSSE.  Voltage angle errors showed the greatest improvement (up to 98\% improvement), and voltage magnitude errors improved substantially, ranging from 61\% to 94\%. The proposed framework reinforces a key component of  EMSs of power systems.

\section{Conclusion}
\label{conc}

This study presented a joint multi-objective optimization framework for sensor placement and attack detection in power systems using PIGTNs. The framework effectively balanced detection performance and practical sensor placement constraints utilizing NSGA-II for robust layout. It optimized the trade-off between detection performance and sensor placement objectives by placing sensors purposefully. 
Extensive evaluations of seven test cases confirmed its consistent detection improvements and robustness to sensor failures,  compared to sensor placement benchmarks.  Its performance gains in TPR and F1-score were achieved without compromising FPR. 
Moreover, integrating AC power flow constraints into the model loss and co-optimizing sensor placement led to strong generalization to unseen LR attacks. The optimized sensor layouts led to an improvement in PSSE performance--up to 98\% reduction in mean state error. Future directions include factoring in  sensor deployment cost, as installation costs vary by location. An adversarial training procedure can be investigated to further improve  the robustness of the detection model to more sophisticated perturbations.

\bibliography{refs}
\bibliographystyle{elsarticle-num}

\end{document}